\documentclass[12pt,twoside]{article}
\usepackage{latexsym,amsmath,amssymb,hyperref}
\usepackage{wrapfig}
  \newdimen\paravsp  \paravsp=1.3ex
\def\beq{\begin{equation}}    \def\eeq{\end{equation}}
\def\beqn{\begin{displaymath}}\def\eeqn{\end{displaymath}}
\def\bqa{\begin{eqnarray}}    \def\eqa{\end{eqnarray}}
\def\bqan{\begin{eqnarray*}}  \def\eqan{\end{eqnarray*}}

\newtheorem{thm}{Theorem}
\newtheorem{cor}[thm]{Corollary}
\newtheorem{exmp}[thm]{Example}
\newtheorem{rem}[thm]{Remark}
\newtheorem{lem}[thm]{Lemma}
\newenvironment{keywords}{\centerline{\bf\small
Keywords}\vspace{-1ex}\begin{quote}\small}{\par\end{quote}\vskip 1ex}
\newcommand{\qed}{\hspace*{\fill}\rule{1.4ex}{1.4ex}$\quad$\\}

\newcommand{\eoe}{\hspace*{\fill} $\diamondsuit\quad$}
\renewcommand{\P}{{\rm P}}
\newcommand{\E}{{\bf E}}
\newcommand{\uP}{{\overline\P}}
\newcommand{\uE}{{\overline\E}}
\newcommand{\lP}{{\underline\P}}
\newcommand{\lE}{{\underline\E}}

\newcommand{\erre}{\mathbf{R}}
\newcommand{\enne}{\mathbf{N}}
\newcommand{\giv}{{\,|\,}}
\newcommand{\ull}{{\Lambda}}
\newcommand{\ul}{{\lambda}}

\newcommand{\ucat}{{\mathcal{X}=\{x_1,\ldots ,x_k\}}}

\newcommand{\ucn}{{\mathcal{X}}^n}
\newcommand{\uc}{{\mathcal{X}}}

\newcommand{\ill}{I}
\newcommand{\healthy}{H}
\newcommand{\postest}{+}
\newcommand{\negtest}{-}

\newcommand{\ux}{{\mathrm{X}}}

\newcommand{\us}{{\mathrm{S}}}

\newcommand{\uxs}{{\mathbf{x}}}

\newcommand{\uss}{{\mathbf{s}}}

\newcommand{\latent}{{\mathbf{X}}}
\newcommand{\latentfut}{{\mathbf{X'}}}
\newcommand{\manifest}{{\mathbf{S}}}
\newcommand{\realized}{{\mathbf{X}}}

\newcommand{\countsx}{{\mathbf{a}=(a_1^{\uxs},\ldots ,a_k^{\uxs})}}

\newcommand{\ut}{{\mathbf{t}}}
\newcommand{\ste}{{\Theta}}

\newcommand{\ssette}{{\Theta:=\{\mathbf{\boldsymbol{\vartheta}}=(\vartheta_1,\ldots,\vartheta_k)\,|\, \sum_{i=1}^k \vartheta
_i=1,\; 0\leq \vartheta_i\leq 1\}}}
\newcommand{\sett}{{\mathcal{T}}}
\newcommand{\ssett}{{\mathcal{T}:=\{\ut=(t_1,\ldots ,t_k)\,|\,
\sum_{j=1}^k\, t_k=1,\, 0<t_j<1\}}}

\newcommand{\seqpn}{{(p_n)_{n\in\enne}}}

\newcommand{\mzero}{{\mathcal{M}_0}}

\begin{document}


\title{\vspace{-4ex}
\vskip 2mm\bf\Large\hrule height5pt \vskip 4mm
Limits of Learning about a Categorical Latent Variable under Prior Near-Ignorance
\vskip 4mm \hrule height2pt}

\author{
{\normalsize\bf Alberto Piatti$^a,$ Marco Zaffalon$^a,$ Fabio Trojani$^b,$ Marcus Hutter$^c$} \\
\scriptsize{\bfseries alberto.piatti@idsia.ch, \ zaffalon@idsia.ch, \ fabio.trojani@unisg.ch, \ marcus@hutter1.net} \\
\small $^a$Istituto Dalle Molle di Studi sull'Intelligenza Artificiale (IDSIA),\\[-1ex]
\small Galleria 2, CH-6928 Manno-Lugano, Switzerland. \\
\small $^b$Institute of Banking and Finance, University of St. Gallen, \\[-1ex]
\small Rosenbergstr. 52, CH-9000 St.Gallen, Switzerland. \\
\small $^c$RSISE, Australian National University and NICTA, \\[-1ex]
\small $^d$ Canberra, ACT, 0200, Australia.
}

\date{April 2009}
\maketitle

\begin{abstract}
In this paper, we consider the coherent theory of (epistemic)
uncertainty of Walley, in which beliefs are represented through sets
of probability distributions, and we focus on the problem of
modeling prior ignorance about a categorical random variable. In
this setting, it is a known result that a state of prior ignorance
is not compatible with learning. To overcome this problem, another
state of beliefs, called \emph{near-ignorance}, has been proposed.
Near-ignorance resembles ignorance very closely, by satisfying some
principles that can arguably be regarded as necessary in a state of
ignorance, and allows learning to take place. What this paper does,
is to provide new and substantial evidence that also near-ignorance
cannot be really regarded as a way out of the problem of starting
statistical inference in conditions of very weak beliefs. The key to
this result is focusing on a setting characterized by a variable of
interest that is \emph{latent}. We argue that such a setting is by
far the most common case in practice, and we provide, for the case
of categorical latent variables (and general \emph{manifest}
variables) a condition that, if satisfied, prevents learning to take
place under prior near-ignorance. This condition is shown to be
easily satisfied even in the most common statistical problems. We
regard these results as a strong form of evidence against the
possibility to adopt a condition of prior near-ignorance in real
statistical problems.
\end{abstract}

\begin{keywords}
Near-ignorance set of priors;
Latent variables;
Imprecise Dirichlet model.
\end{keywords}

\newpage
\section{Introduction}\label{sec:introduction}

Epistemic theories of statistics are often confronted with the
question of \emph{prior ignorance}. Prior ignorance means that a
subject, who is about to perform a statistical analysis, is missing
substantial beliefs about the underlying data-generating process.
Yet, the subject would like to exploit the available sample to draw
some statistical conclusion, i.e., the subject would like to use the
data to learn, moving away from the initial condition of ignorance.
This situation is very important as it is often desirable to start a
statistical analysis with weak assumptions about the problem of
interest, thus trying to implement an objective-minded approach to
statistics.

A fundamental question is whether prior ignorance is compatible with
learning or not. Walley gives a negative answer for the case of his
self-consistent (or \emph{coherent}) theory of statistics based on
the modeling of beliefs through sets of probability distributions.
He shows, in a very general sense, that \emph{vacuous} prior
beliefs, i.e., beliefs that a priori are maximally imprecise, lead
to vacuous posterior beliefs, irrespective of the type and amount of
observed data \cite[Section~7.3.7]{Walley1991}. At the same time, he
proposes focusing on a slightly different state of beliefs, called
\emph{near-ignorance}, that does enable learning to take place
\cite[Section~4.6.9]{Walley1991}. Loosely speaking, near-ignorant
beliefs are beliefs that are vacuous for a proper subset of the
functions of the random variables under consideration (see
Section~\ref{sec:nearignorance}). In this way, a near-ignorance
prior still gives one the possibility to express vacuous beliefs for
some functions of interest, and at the same time it maintains the
possibility to learn from data. The fact that learning is possible
under prior near-ignorance is shown, for instance, in the special
case of the \emph{imprecise Dirichlet model} (IDM)
\cite{Walley1996,Bernard2005}. This is a popular model, based on a
near-ignorance set of priors, used in the case of inference from
categorical data generated by a multinomial process.

Our aim in this paper is to investigate whether near-ignorance can
be really regarded as a possible way out of the problem of starting
statistical inference in conditions of very weak beliefs. We carry
out this investigation in a setting made of categorical data
generated by a multinomial process, like in the IDM, but we consider
near-ignorance sets of priors in general, not only that used in the
IDM.

The interest in this investigation is motivated by the fact that
near-ignorance sets of priors appear to play a crucially important
role in the question of modeling prior ignorance about a categorical
random variable. The key point is that near-ignorance sets of priors
can be made to satisfy two principles: the \emph{symmetry} and the
\emph{embedding principles}. The first is well known and is
equivalent to Laplace's \emph{indifference principle}; the second
states, loosely speaking, that if we are ignorant a priori, our
prior beliefs on an event of interest should not depend on the space
of possibilities in which the event is embedded (see
Section~\ref{sec:nearignorance} for a discussion about these two
principles). Walley \cite{Walley1991}, and later de Cooman and
Miranda \cite{DeCooman2006}, have argued extensively on the
necessity of both the symmetry and the embedding principles in order
to characterize a condition of ignorance about a categorical random
variable. This implies, if we agree that the symmetry and the
embedding principles are necessary for ignorance, that
near-ignorance sets of priors should be regarded as an especially
important avenue for a subject who wishes to learn starting in a
condition of ignorance.

Our investigation starts by focusing on a setting where the
categorical variable $\ux$ under consideration is \emph{latent}.
This means that we cannot observe the realizations of $\ux$, so that
we can learn about it only by means of another, not necessarily
categorical, variable $\us$, related to $\ux$ through a known
conditional probability distribution $P(\us\giv\ux)$. Variable $\us$
is assumed to be \emph{manifest}, in the sense that its realizations
can be observed (see Section~\ref{sec:latent}). The intuition behind
the setup considered, made of $\ux$ and $\us$, is that in many real
cases it is not possible to directly observe the value of a random
variable in which we are interested, for instance when this variable
represents a patient's health and we are observing the result of a
diagnostic test. In these cases, we need to use a manifest variable
(the medical test) in order to obtain information about the original
latent variable (the patient's health). In this paper, we regard the
passage from the latent to the manifest variable as made by a
process that we call the \emph{observational
process}.\footnote{Elsewhere, this is also called the
\emph{measurement process}.}

Using the introduced setup, we give a condition in
Section~\ref{sec:conditions}, related to the likelihood function,
that is shown to be sufficient to prevent learning about $\ux$ under
prior near-ignorance. The condition is very general as it is
developed for any set of priors that models near-ignorance (thus
including the case of the IDM), and for very general kinds of
probabilistic relations between $\ux$ and $\us$. We show then, by
simple examples, that such a condition is easily satisfied, even in
the most elementary and common statistical problems.

In order to fully appreciate this result, it is important to realize
that latent variables are ubiquitous in problems of uncertainty. The
key point here is that the scope of observational processes greatly
extends if we consider that even when we directly obtain the value
of a variable of interest, what we actually obtain is the
observation of the value rather than the value itself. Doing this
distinction makes sense because in practice an observational process
is usually imperfect, i.e., there is very often (it could be argued
that there is always) a positive probability of confounding the
realized value of $\ux$ with another possible value committing thus
an observation error.

Of course, if the probability of an observation error is very small
and we consider one of the common Bayesian model proposed to learn
under prior ignorance, then there is little difference between the
results provided by a latent variable model modeling correctly the
observational process, and the results provided by a model where the
observations are assumed to be perfect. For this reason, the
observational process is often neglected in practice and the
distinction between the latent variable and the manifest one is not
enforced.

But, on the other hand, if we consider sets of probability
distributions to model our prior beliefs, instead of a single
probability distribution, and in particular if we consider
near-ignorance sets of priors, then there can be an extreme
difference between a latent variable model and a model where the
observations are considered to be perfect, so that learning may be
impossible in the first model and possible in the second. As a
consequence, when dealing with sets of probability distributions,
neglecting the observational process may be no longer justified even
if the probability of observation error is tiny. This is shown in a
definite sense in Example~\ref{exmp:diagnostictest2} of
Section~\ref{sec:conditions,subsec:IDM}, where we analyze the
relevance of our results for the special case of the IDM. From the
proofs in this paper, it follows that this kind of behavior is
mainly determined by the presence, in the near-ignorance set of
priors, of extreme, almost-deterministic, distributions. And the
question is that these problematic distributions, which are usually
not considered when dealing with Bayesian models with a single
prior, cannot be ruled out without dropping near-ignorance.

These considerations highlight the quite general applicability of
the present results and raise hence serious doubts about the
possibility to adopt a condition of prior near-ignorance in real, as
opposed to idealized, applications of statistics. As a consequence,
it may make sense to consider re-focusing the research about this
subject on developing models of very weak states of belief that are,
however, stronger than near-ignorance. This might also involve
dropping the idea that both the symmetry and the embedding
principles can be realistically met in practice.

\section{Categorical Latent Variables}\label{sec:latent}

In this paper, we follow the general definition of \emph{latent} and
\emph{manifest variables} given by Skrondal and Rabe-Hasketh
\cite{Skrondal2004}: a \emph{latent variable} is a random variable
whose realizations are unobservable (hidden), while a \emph{manifest
variable} is a random variable whose realizations can be directly
observed.

The concept of latent variable is central in many sciences, like for
example psychology and medicine. Skrondal and Rabe-Hasketh list
several fields of application and several phenomena that can be
modelled using latent variables, and conclude that latent variable
modeling ``\emph{pervades modern mainstream statistics},'' although
``\emph{this omni-presence of latent variables is commonly not
recognized, perhaps because latent variables are given different
names in different literatures, such as random effects, common
factors and latent classes},'' or hidden variables.

But what are latent variables in practice? According to Boorsbom et
al. \cite{Boorsbom2002}, there may be different interpretations of
latent variables. A latent variable can be regarded, for example, as
an unobservable random variable that exists independently of the
observation. An example is the unobservable health status of a
patient that is subject to a medical test. Another possibility is to
regard a latent variable as a product of the human mind, a construct
that does not exist independently of the observation. For example
the \emph{unobservable state of the economy}, often used in economic
models. In this paper, we assume the existence of a latent
categorical random variable $\ux$, with outcomes in $\ucat$ and
unknown chances $\boldsymbol{\vartheta}\in\ssette$, without
stressing any particular interpretation. Throughout the paper, we
denote by $\boldsymbol{\vartheta}$ a particular vector of chances in
$\ste$ and by $\boldsymbol{\theta}$ a random variable on $\ste$.

Now, let us focus on a bounded real-valued function $f$ defined on
$\ste$, where $\boldsymbol{\vartheta}\in\ste$ are the unknown
chances of $\ux$. We aim at learning the value
$f(\boldsymbol{\vartheta})$ using $n$ realizations of the variable
$\ux$. Because the variable $\ux$ is latent and therefore
unobservable by definition, the only way to learn
$f(\boldsymbol{\vartheta})$ is to observe the realizations of some
manifest variable $\us$ related, through known probabilities
$P(\us\giv\ux)$, to the (unobservable) realizations of $\ux$. An
example of known probabilistic relationship between latent and
manifest variables is the following.

\begin{exmp}\rm
\label{exmp:introtest} Consider a binary medical diagnostic test
used to assess the health status of a patient with respect to a
given disease. The accuracy of a diagnostic test\footnote{For
further details about the modeling of diagnostic accuracy with
latent variables see Yang and Becker \cite{Yang1997}.} is determined
by two probabilities: the \emph{sensitivity} of a test is the
probability of obtaining a positive result if the patient is
diseased; the \emph{specificity} is the probability of obtaining a
negative result if the patient is healthy. Medical tests are assumed
to be imperfect indicators of the unobservable true disease status
of the patient. Therefore, we assume that the probability of
obtaining a positive result when the patient is healthy,
respectively of obtaining a negative result if the patient is
diseased, are non-zero. Suppose, to make things simpler, that the
sensitivity and the specificity of the test are known. In this
example, the unobservable health status of the patient can be
considered as a binary latent variable $\ux$ with values in the set
$\{\textrm{Healthy},\textrm{Ill}\}$, while the result of the test
can be considered as a binary manifest variable $\us$ with values in
the set $\{\textrm{Negative result},\textrm{Positive result}\}$.
Because the sensitivity and the specificity of the test are known,
we know $P(\us\giv\ux)$.\eoe
\end{exmp}

We continue discussion about this example later on, in the light of
our results, in Example~\ref{exmp:diagnostictest} of
Section~\ref{sec:conditions}.

\section{Near-ignorance sets of priors}\label{sec:nearignorance}

Consider a categorical random variable $\ux$ with outcomes in
$\ucat$ and unknown chances $\boldsymbol{\vartheta}\in\ste$. Suppose
that we have no relevant prior information about
$\boldsymbol{\vartheta}$ and we are therefore in a situation of
prior ignorance about $\ux$. How should we model our prior beliefs
in order to reflect the initial lack of knowledge?

Let us give a brief overview of this topic in the case of coherent
models of uncertainty, such as Bayesian probability theory and Walley's
theory of \emph{coherent lower previsions}.

In the traditional Bayesian setting, prior beliefs are modelled
using a single prior probability distribution. The problem of
defining a standard prior probability distribution modeling a
situation of prior ignorance, a so-called \emph{non-informative
prior}, has been an important research topic in the last two
centuries\footnote{Starting from the work of Laplace at the
beginning of the 19$^\text{th}$ century \cite{Laplace1820}.} and,
despite the numerous contributions, it remains an open research
issue, as illustrated by Kass and Wassermann \cite{Kass1996}. See
also Hutter \cite{Hutter2006} for recent developments and
complementary considerations. There are many principles and
properties that are desirable when the focus is on modeling a
situation of prior ignorance, and that have indeed been used in past
research to define non-informative priors. For example Laplace's
\emph{symmetry or indifference} principle has suggested, in case of
finite possibility spaces, the use of the uniform distribution.
Other principles, like for example the principle of \emph{invariance
under group transformations}, the \emph{maximum entropy} principle,
the \emph{conjugate priors} principle, etc., have suggested the use
of other non-informative priors, in particular for continuous
possibility spaces, satisfying one or more of these principles. But,
in general, it has proven to be difficult to define a standard
non-informative prior satisfying, at the same time, all the
desirable principles.

We follow Walley \cite{Walley1996} and de Cooman and Miranda
\cite{DeCooman2006} when they say that there are at least two
principles that should be satisfied to model a situation of prior
ignorance: the \emph{symmetry} and the \emph{embedding principles}.
The \emph{symmetry principle} states that, if we are ignorant a
priori about $\boldsymbol{\vartheta}$, then we have no reason to
favour one possible outcome of $\ux$ over another, and therefore our
probability model on $\ux$ should be symmetric. This principle is
equivalent to Laplace's \emph{symmetry or indifference} principle.
The \emph{embedding principle} states that, for each possible event
$A$, the probability assigned to $A$ should not depend on the
possibility space $\uc$ in which $A$ is embedded. In particular, the
probability assigned a priori to the event $A$ should be invariant
with respect to refinements and coarsenings of $\uc$.

It is easy to show that the embedding principle is not satisfied by
the uniform distribution. How should we model our prior ignorance in
order to satisfy these two principles? Walley\footnote{In Walley
\cite{Walley1991}, Note~7 at p.~526. See also Section~5.5 of the
same book.} gives what we believe to be a compelling answer to this
question: he proves that the only coherent probability model on
$\ux$ consistent with the two principles is the \emph{vacuous
probability model}, i.e., the model that assigns, for each
non-trivial event $A$, lower probability $\lP(A)=0$ and upper
probability $\uP(A)=1$. Clearly, the vacuous probability model
cannot be expressed using a single probability distribution. It
follows then, if we agree that the symmetry and the embedding
principles are characteristics of prior ignorance, that we need
\emph{imprecise probabilities} to model such a state of
beliefs.\footnote{For a complementary point of view, see Hutter
\cite{Hutter2006}.} Unfortunately, it is easy to show that updating
the vacuous probability model on $\ux$ produces only vacuous
posterior probabilities. Therefore, the vacuous probability model
alone is not a viable way to address our initial problem. Walley
suggests, as an alternative, the use of \emph{near-ignorance sets of
priors}.\footnote{Walley calls a set of probability distributions
modeling near-ignorance a \emph{near-ignorance prior}. In this paper
we use the term \emph{near-ignorance set of priors} in order to
avoid confusion with the precise Bayesian case.}

A near-ignorance set of priors is a probability model on the chances
$\boldsymbol{\theta}$ of $\ux$, modeling a very weak state of
knowledge about $\boldsymbol{\theta}$. In practice, a near-ignorance
set of priors is a large closed convex set $\mzero$ of prior
probability densities on $\boldsymbol{\theta}$ which produces
\emph{vacuous expectation}s for various but not all functions $f$ on
$\ste$, i.e., such that
$\lE(f)=\inf_{\boldsymbol{\vartheta}\in\Theta}
f(\boldsymbol{\vartheta})$ and
$\uE(f)=\sup_{\boldsymbol{\vartheta}\in\Theta}
f(\boldsymbol{\vartheta})$.

The key point here is that near-ignorance sets of priors can be
designed so as to satisfy both the symmetry and the embedding
principles. In fact, if a near-ignorance set of priors produces
vacuous expectations for all the functions
$f(\boldsymbol{\vartheta})=\vartheta_i$ for each
$i\in\{1,\ldots,k\}$, then, because a priori
$P(\ux=x_i)=E(\theta_i)$, the near-ignorance set of priors implies
the vacuous probability model on $\ux$ and satisfies therefore both
the symmetry and the embedding principle, thus delivering a
satisfactory model of prior near-ignorance.\footnote{We call this
state near-ignorance because, although we are completely ignorant a
priori about $\ux$, we are not completely ignorant about
$\boldsymbol{\theta}$ \cite[Section~5.3,~Note~4]{Walley1991}.}
Updating a near-ignorance prior consists in updating all the
probability densities in $\mzero$ using Bayes' rule. Since the
beliefs on $\boldsymbol{\theta}$ are not vacuous, this makes it
possible to calculate non-vacuous posterior probabilities for $\ux$.

A good example of near-ignorance set of priors is the set $\mzero$
used in the \emph{imprecise Dirichlet model}. The IDM models a
situation of prior near-ignorance about a categorical random
variable $\ux$. The near-ignorance set of priors $\mzero$ used in
the IDM consists of the set of all Dirichlet
densities\footnote{Throughout the paper, if no confusion is
possible, we denote the outcome
$\boldsymbol{\theta}=\boldsymbol{\vartheta}$ by
$\boldsymbol{\vartheta}$. For example, we denote
$p(\boldsymbol{\theta}=\boldsymbol{\vartheta})$ by
$p(\boldsymbol{\vartheta})$.}
$p(\boldsymbol{\vartheta})=dir_{s,\ut}(\boldsymbol{\vartheta})$ for
a fixed $s>0$ and all $\ut\in\sett$, where
\beq\label{eq:dir}
  dir_{s,\ut}(\boldsymbol{\vartheta}):=\frac{\Gamma (s)}{\prod_{i=1}^k
  \Gamma (st_i)}\,\prod_{i=1}^k\, \vartheta_i^{st_i-1},
\eeq
and
\beq\label{eq:setT}
  \ssett.
\eeq
The particular choice of $\mzero$ in the IDM implies vacuous prior
expectations for all the functions
$f(\boldsymbol{\vartheta})=\vartheta_i^{R}$, for all integers $R\geq
1$ and all $i\in\{1,\ldots ,k\}$, i.e., $\lE(\theta_i^{R})=0$ and
$\uE(\theta_i^{R})=1$. Choosing $R=1$, we have, a priori,
\beqn
  \lP(\ux=x_i)=\lE(\theta_i)=0,\quad
  \uP(\ux=x_i)=\uE(\theta_i)=1.
\eeqn
It follows that the particular near-ignorance set of priors $\mzero$
used in the IDM implies a priori the vacuous probability model on
$\ux$ and, therefore, satisfies both the symmetry and embedding
principles. On the other hand, the particular set of priors used in
the IDM does not imply vacuous prior expectations for all the
functions $f(\boldsymbol{\vartheta})$. For example, vacuous
expectations for the functions
$f(\boldsymbol{\vartheta})=\vartheta_i\cdot\vartheta_j$ for $i\neq
j$ would be $\lE(\vartheta_i\cdot\vartheta_j)=0$ and
$\uE(\vartheta_i\cdot\vartheta_j)=0.25$, but in the IDM we have a
priori $\uE(\vartheta_i\cdot\vartheta_j)<0.25$ and the prior
expectations are therefore not vacuous. In Walley \cite{Walley1996},
it is shown that the IDM produces, for each observed dataset,
non-vacuous posterior probabilities for $\ux$.

\section{Limits of Learning under Prior Near-Ignorance}\label{sec:conditions}

Consider a sequence of independent and identically distributed (IID)
categorical latent variables $(\ux_i)_{i\in\enne}$ with outcomes in
$\uc$ and unknown chances $\boldsymbol{\theta}$, and a sequence of
independent manifest variables $(\us_i)_{i\in\enne}$, which we allow
to be defined either on finite or infinite spaces. We assume that a
realization of the manifest variable $\us_i$ can be observed only
after a (hidden) realization of the latent variable $\ux_i$.
Furthermore, we assume $\us_i$ to be independent of the chances
$\boldsymbol{\theta}$ of $\ux_i$ conditional on $\ux_i$, i.e.,
\beq\label{eq:independence}
  P(\us_i\giv \ux_i=x_j,\boldsymbol{\theta}=\boldsymbol{\vartheta})=P(\us_i\giv \ux_i=x_j),
\eeq
for each $x_j\in\uc$ and
$\boldsymbol{\vartheta}\in\ste$.\footnote{We denote usually by $P$ a
probability (discrete case) and with $p$ a probability density
(continuous case). If an expression holds in both the discrete and
the continuous case, like for example Equation
(\ref{eq:independence}), then we use $P$ to indicate both cases.}
These assumptions model a two-step process where the variable
$\us_i$ is used to convey information about the realized value of
$\ux_i$ for each $i$, independently of the chances of $\ux_i$. The
(in)dependence structure can be depicted graphically as follows,

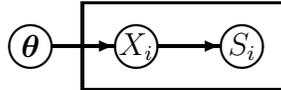
\begin{center}
\unitlength=0.8pt
\begin{picture}(130,45)(10,5)
\thicklines \put(20,30){\circle{20}\makebox(0,0)[cc]{$\boldsymbol{\theta}$}}
\put(70,30){\circle{20}\makebox(0,0)[cc]{$X_i$}}
\put(120,30){\circle{20}\makebox(0,0)[cc]{$S_i$}}
\put(45,10){\framebox(95,40)}
\put(30,30){\vector(1,0){30}} \put(80,30){\vector(1,0){30}}
\end{picture}
\end{center}

where the framed part of this structure is what we call an
\emph{observational process}.

To make things simpler, we assume the probability distribution
$P(\us_i\giv \ux_i=x_j)$ to be precise and known for each
$x_j\in\uc$ and each $i\in\enne$.

We divide the discussion about the limits of learning under prior
near-ignorance in three subsections. In
Section~\ref{sec:conditions,subsec:generalcase} we discuss our
general parametric problem and we obtain a condition that, if
satisfied, prevents learning to take place. In
Section~\ref{sec:conditions,subsec:predictiveprob} we study the
consequences of our theoretical results in the particular case of
predictive probabilities. Finally, in
Section~\ref{sec:conditions,subsec:IDM}, we focus on the particular
near-ignorance set of priors used in the IDM and we obtain necessary
and sufficient conditions for learning with categorical manifest
variables.

\subsection{General parametric inference}\label{sec:conditions,subsec:generalcase}
We focus on a very general problem of parametric inference. Suppose
that we observe a dataset $\uss$ of realizations of the manifest
variables $\us_1,\ldots,\us_n$ related to the (unobservable) dataset
$\uxs\in\uc^n$ of realizations of the variables
$\ux_1,\ldots,\ux_n$. Defining the random variables
$\latent:=(\ux_1,\ldots,\ux_n)$ and
$\manifest:=(\us_1,\ldots,\us_n)$, we have $\manifest=\uss$ and
$\latent=\uxs$. To simplify notation, when no confusion can arise,
we denote in the rest of the paper $\manifest=\uss$ with $\uss$.
Given a bounded function $f(\boldsymbol{\vartheta})$, our aim is to
calculate $\lE(f\giv\uss)$ and $\uE(f\giv\uss)$ starting from a
condition of ignorance about $f$, i.e., using a near ignorance prior
$\mzero$, such that
$\lE(f)=f_{\min}:=\inf_{\boldsymbol{\vartheta}\in\Theta}
f(\boldsymbol{\vartheta})$ and
$\uE(f)=f_{\max}:=\sup_{\boldsymbol{\vartheta}\in\Theta}
f(\boldsymbol{\vartheta})$.

Is it really possible to learn something about the function $f$,
starting from a condition of prior near-ignorance and having
observed a dataset $\uss$? The following theorem shows that, very
often, this is not the case. In particular,
Corollary~\ref{cor:fondamentale} shows that there is a condition
that, if satisfied, prevents learning to take place.

\begin{thm}\label{thm:fondamentale}
Let $\uss$ be given. Consider a bounded continuous function $f$
defined on $\Theta$ and a near-ignorance set of priors $\mzero$. Then the
following statements hold.\footnote{The proof of this theorem is
given in the appendix, together with all the other proofs of the
paper.}
\begin{enumerate}
\item If the likelihood function $\P(\uss\giv\boldsymbol{\vartheta})$ is strictly
positive\footnote{In the appendix it is shown that the assumptions
of positivity of $\P(\uss\giv\boldsymbol{\vartheta})$ in
Theorem~\ref{thm:fondamentale} can be substituted by the following
weaker assumptions. For a given arbitrary small $\delta>0$, denote by $\Theta_{\delta}$ the measurable set,
$\Theta_{\delta}:=\{\boldsymbol{\vartheta}\in\Theta\,|\, f(\boldsymbol{\vartheta})\geq
f_{\max}-\delta\}.$ If $\P(\uss\giv\boldsymbol{\vartheta})$ is such that, $
\lim_{\delta\rightarrow
0}\inf_{\boldsymbol{\vartheta}\in\Theta_{\delta}}\P(\uss\giv\boldsymbol{\vartheta})=c>0, $
then Statement 1 of Theorem ~\ref{thm:fondamentale} holds. The same
holds for the second statement, substituting $\ste_{\delta}$ with
$\widetilde{\Theta}_{\delta}:=\{\boldsymbol{\vartheta}\in\Theta\,|\,
f(\boldsymbol{\vartheta})\leq f_{\min}+\delta\}.$} in each point in which $f$
reaches its maximum value $f_{\max}$, is continuous in an arbitrary
small neighborhood of those points, and $\mzero$ is such that a
priori $\uE(f)=f_{\max}$, then
\beqn
  \uE(f\giv\uss)=\uE(f)=f_{\max}.
\eeqn
\item If the likelihood function $\P(\uss\giv\boldsymbol{\vartheta})$ is strictly
positive in each point in which $f$ reaches its minimum value
$f_{\min}$, is continuous in an arbitrary small neighborhood of
those points, and $\mzero$ is such that a priori $\lE(f)=f_{\min}$,
then
\beqn
  \lE(f\giv\uss)=\lE(f)=f_{\min}.
\eeqn
\end{enumerate}
\end{thm}

\begin{cor}\label{cor:fondamentale}
Consider a near-ignorance set of priors $\mzero$. Let $\uss$ be given and
let $\P(\uss\giv\boldsymbol{\vartheta})$ be a continuous strictly positive function
on $\Theta$. If $\mzero$ is such that $\lE(f)=f_{\min}$ and
$\uE(f)=f_{\max}$, then
\bqan
  \lE(f\giv\uss) &=& \lE(f)=f_{\min},\\
  \uE(f\giv\uss) &=& \uE(f)=f_{\max}.
\eqan
\end{cor}
In other words, given $\uss$, if the likelihood function is strictly
positive, then the functions $f$ that, according to $\mzero$, have
vacuous expectations a priori, have vacuous expectations also a
posteriori, after having observed $\uss$. It follows that, if this
sufficient condition is satisfied, we cannot use near-ignorance
priors to model a state of prior ignorance because only vacuous
posterior expectations are produced. The sufficient condition
described above is met very easily in practice, as shown in the
following two examples. In the first example, we consider a very
simple setting where the manifest variables are categorical. In the
second example, we consider a simple setting with continuous
manifest variables. We show that, in both cases, the sufficient
condition is satisfied and therefore we are unable to learn under
prior near-ignorance.

\begin{exmp}\label{exmp:diagnostictest}\rm
Consider the medical test introduced in Example~\ref{exmp:introtest}
and an (ideally) infinite population of individuals. Denote by the
binary variable $\ux_i\in\{\healthy,\ill\}$ the health status of the
$i$-th individual of the population and with
$\us_i\in\{\postest,\negtest\}$ the results of the diagnostic test
applied to the same individual. We assume that the variables in the
sequence $(\ux_i)_{i\in\enne}$ are IID with unknown chances
$(\vartheta,1-\vartheta)$, where $\vartheta$ corresponds to the
(unknown) proportion of diseased individuals in the population.
Denote by $1-\varepsilon_1$ the specificity and with
$1-\varepsilon_2$ the sensitivity of the test. Then it holds that
\beqn
\P(\us_i=\postest\giv\ux_i=\healthy)=\varepsilon_1>0, \quad
\P(\us_i=\negtest\giv\ux_i=\ill)=\varepsilon_2>0,
\eeqn
where ($\ill,\healthy,\postest,\negtest$) denote (patient ill,
patient healthy, test positive, test negative).

Suppose that we observe the results of the test applied to $n$
different individuals of the population; using our previous notation
we have $\manifest=\uss$. For each individual we have,
\begin{align*}
& \P(\us_i=\postest\giv\vartheta)\\=&\P(\us_i=\postest\giv\ux_i=\ill)
  \P(\ux_i=\ill\giv\vartheta)+\P(\us_i=\postest\giv\ux_i=\healthy)\P(\ux_i=\healthy\giv\vartheta)\\
=& \underbrace{(1-\varepsilon_2)}_{>0}\cdot\vartheta+\underbrace{\varepsilon_1}_{>0}\cdot (1-\vartheta)>0.\\
\end{align*}
Analogously,
\begin{align*}
& \P(\us_i=\negtest\giv\vartheta)\\
= & \P(\us_i=\negtest\giv\ux_i=\ill)\P(\ux_i=\ill\giv\vartheta)+\P(\us_i=\negtest\giv\ux_i=\healthy)\P(\ux_i=\healthy\giv\vartheta)\\
= & \underbrace{\varepsilon_2}_{>0}\cdot\vartheta+\underbrace{(1-\varepsilon_1)}_{>0}\cdot (1-\vartheta)>0.\\
\end{align*}
Denote by $n^{\uss}$ the number of positive tests in the observed
sample $\uss$. Since the variables $\us_i$ are independent,
we have
\begin{align*}
&
\P(\manifest=\uss\giv\vartheta)=((1-\varepsilon_2)\cdot\vartheta+\varepsilon_1\cdot
(1-\vartheta))^{n^{\uss}}\cdot (\varepsilon_2\cdot\vartheta+(1-\varepsilon_1)\cdot (1-\vartheta))^{n-n^{\uss}}>0\\
\end{align*}
for each $\vartheta\in [0,1]$ and each $\uss\in\uc^n$. Therefore,
according to Corollary~\ref{cor:fondamentale}, all the functions $f$
that, according to $\mzero$, have vacuous expectations a priori have
vacuous expectations also a posteriori. It follows that, if we want
to avoid vacuous posterior expectations, then we cannot model our
prior knowledge (ignorance) using a near-ignorance set of priors.
This simple example shows that our previous theoretical results
raise serious questions about the use of near-ignorance sets of
priors also in very simple, common, and important situations.\eoe
\end{exmp}

Example~\ref{exmp:diagnostictest} focuses on categorical latent and
manifest variables. In the next example, we show that our
theoretical results have important implications also in models with
categorical latent variables and continuous manifest variables.

\begin{exmp}\label{exmp:continuousmanifest}\rm
Consider a sequence of IID categorical variables
$(\ux_i)_{i\in\enne}$ with outcomes in $\uc^n$ and unknown chances
$\boldsymbol{\theta}\in\Theta$. Suppose that, for each $i\geq 1$,
after a realization of the latent variable $\ux_i$, we can observe a
realization of a continuous manifest variable $\us_i$. Assume that
$p(\us_i\giv\ux_i=x_j)$ is a continuous positive probability
density, e.g., a normal $N(\mu_j,\sigma_j^2)$ density, for each
$x_j\in\uc$. We have
\begin{align*}
p(\us_i\giv\boldsymbol{\vartheta}) & = \sum_{x_j\in\uc} p(\us_i\giv\ux_i=x_j)\cdot
\P(\ux_i=x_j\giv\boldsymbol{\vartheta})=\sum_{x_j\in\uc}
\underbrace{p(\us_i\giv\ux_i=x_j)}_{>0}\cdot \vartheta_j>0,
\end{align*}
because $\vartheta_j$ is positive for at least one $j\in\{1,\ldots
,k\}$ and we have assumed $\us_i$ to be independent of $\boldsymbol{\theta}$
given $\ux_i$. Because we have assumed $(\us_i)_{i\in\enne}$ to be a
sequence of independent variables, we have
\beqn
  p(\manifest=\uss\giv\boldsymbol{\vartheta})=\prod_{i=1}^n \underbrace{p(\us_i=\uss_i\giv\boldsymbol{\vartheta})}_{>0}>0.
\eeqn
Therefore, according to Corollary~\ref{cor:fondamentale}, if we
model our prior knowledge using a near-ignorance set of priors
$\mzero$, the vacuous prior expectations implied by $\mzero$ remain
vacuous a posteriori. It follows that, if we want to avoid vacuous
posterior expectations, we cannot model our prior knowledge using a
near-ignorance set of priors.\eoe
\end{exmp}

Examples~\ref{exmp:diagnostictest} and~\ref{exmp:continuousmanifest}
raise, in general, serious criticisms about the use of
near-ignorance sets of priors in real applications.

\subsection{An important special case: predictive probabilities}\label{sec:conditions,subsec:predictiveprob}
We focus now on a very important special case: that of predictive
inference.\footnote{For a general presentation of predictive
inference see Geisser \cite{Geisser1993}; for a discussion of the
imprecise probability approach to predictive inference see Walley
and Bernard \cite{Walley1999}.} Suppose that our aim is to predict
the outcomes of the next $n'$ variables
$\ux_{n+1},\ldots,\ux_{n+n'}$. Let
$\latentfut:=(\ux_{n+1},\ldots,\ux_{n+n'})$. If no confusion is
possible, we denote $\latentfut=\uxs'$ by $\uxs'$. Given
$\uxs'\in\uc^{n'}$, our aim is to calculate $\lP(\uxs'\giv\uss)$ and
$\uP(\uxs'\giv\uss)$. Modeling our prior ignorance about the
parameters $\boldsymbol{\theta}$ with a near-ignorance set of priors
$\mzero$ and denoting by $\mathbf{n'}:=(n_1',\ldots,n_k')$ the
frequencies of the dataset $\uxs'$, we have
\bqan
  \lP(\mathbf{\uxs'}\giv\uss) & = &
  \inf_{p\in\mzero}\P_p(\mathbf{\uxs'}\giv\uss) =
  \inf_{p\in\mzero}\int_{\ste} \prod_{i=1}^k \vartheta_i^{n_i'}
  p(\boldsymbol{\vartheta}\giv\uss) d\boldsymbol{\vartheta}=\\\nonumber & = &
  \inf_{p\in\mzero}\E_p\left(\prod_{i=1}^k
  \vartheta_i^{n_i'}\giv\uss\right) = \lE\left(\prod_{i=1}^k
  \vartheta_i^{n_i'}\giv\uss\right),\label{eq:lowerexpected}\\
\eqan
where, according to Bayes' rule,
\beqn
  p(\boldsymbol{\vartheta}\giv\uss)=
  \frac{\P(\uss\giv\boldsymbol{\vartheta})
  p(\boldsymbol{\vartheta})}{\int_{\ste}
  \P(\uss\giv\boldsymbol{\vartheta}) p(\boldsymbol{\vartheta})
  d\boldsymbol{\vartheta}},
\eeqn
provided that $\int_{\ste} \P(\uss\giv\boldsymbol{\vartheta})
p(\boldsymbol{\vartheta}) d\boldsymbol{\vartheta}\neq 0$.
Analogously, substituting $\sup$ to $\inf$ in
(\ref{eq:lowerexpected}), we obtain
\beq\label{eq:upperexpected}
  \uP(\mathbf{\uxs'}\giv\uss)=\uE\left(\prod_{i=1}^k
  \vartheta_i^{n_i'}\giv\uss\right).
\eeq
Therefore, the lower and upper probabilities assigned to the dataset
$\uxs'$ a priori (a posteriori) correspond to the prior (posterior)
lower and upper expectations of the continuous bounded function
$f(\boldsymbol{\vartheta})=\prod_{i=1}^k \vartheta_i^{n_i'}$.

It is easy to show that, in this case, the minimum of $f$ is 0 and
is reached in all the points $\boldsymbol{\vartheta}\in\ste$ with
$\vartheta_i=0$ for some $i$ such that $n_i'>0$, while the maximum
of $f$ is reached in a single point of $\ste$ corresponding to the
relative frequencies $\mathbf{f'}$ of the sample $\uxs'$, i.e., at
$\mathbf{f'}=\left(\frac{n_1'}{n'},\ldots,\frac{n_k'}{n'}\right)\in\Theta$,
and the maximum of $f$ is given by $\prod_{i=1}^k
\left(\frac{n_i'}{n'}\right)^{n_i'}$. It follows that the maximally
imprecise probabilities regarding the dataset $\uxs'$, given that
$\uxs'$ has been generated by a multinomial process, are given by
\beqn
  \lP(\uxs')=\lE\left(\prod_{i=1}^k \theta_i^{n_i'}\right)=0,\quad
  \uP(\uxs')=\uE\left(\prod_{i=1}^k \theta_i^{n_i'}\right)=\prod_{i=1}^k
\left(\frac{n_i'}{n'}\right)^{n_i'}.
\eeqn
The general results stated in
Section~\ref{sec:conditions,subsec:generalcase} hold also in the
particular case of predictive probabilities. In particular,
Corollary~\ref{cor:fondamentale} can be rewritten as follows.
\begin{cor}\label{cor:fondamentale4}
Consider a near-ignorance set of priors $\mzero$. Let $\uss$ be
given and let $\P(\uss\giv\boldsymbol{\vartheta})$ be a continuous
strictly positive function on $\Theta$. Then, if $\mzero$ implies
prior probabilities for a dataset $\uxs'\in\uc^{n'}$ that are
maximally imprecise, the predictive probabilities of $\uxs'$ are
maximally imprecise also a posteriori, after having observed $\uss$,
i.e.,
\beqn
  \lP(\uxs'\giv\uss)=\lP(\uxs')=0,\quad \uP(\uxs'\giv\uss)=\uP(\uxs') =\prod_{i=1}^k
  \left(\frac{n_i'}{n'}\right)^{n_i'}.
\eeqn
\end{cor}
%
\subsection{Predicting the next outcome with categorical manifest variables}\label{sec:conditions,subsec:IDM}
In this section we consider a special case for which we give
necessary and sufficient conditions to learn under prior
near-ignorance. These conditions are then used to analyze the IDM.

We assume that all the manifest variables in $\manifest$ are
categorical. Given an arbitrary categorical manifest variable
$\us_i$, denote by $\mathcal{S}^i:=\{s_1,\ldots ,s_{n_i}\}$ the
finite set of possible outcomes of $\us_i$. The probabilities of
$\us_i$ are defined conditional on the realized value of $\ux_i$ and
are given by
\beqn
  \ul_{hj}(\us_i):=P(\us_i=s_h\giv \ux_i=x_j),
\eeqn
where $h\in\{1,\ldots ,n_i\}$ and $j\in\{1,\ldots ,k\}$. The
probabilities of $\us_i$ can be collected in a $n_i\times k$
stochastic matrix $\ull^{\us_i}$ defined by
\beqn
  \ull(\us_i):=\left(\begin{array}{ccc} \ul_{11}(\us_i) & \ldots & \ul_{1k}(\us_i)\\
  \vdots & \ddots & \vdots\\
  \ul_{n_i1}(\us_i) & \ldots &
  \ul_{n_ik}(\us_i)\\\end{array}\right),
\eeqn
which is called \emph{emission matrix} of $\us_i$.

Our aim, given $\uss$, is to predict the next (latent) outcome
starting from prior near-ignorance. In other words, our aim is to
calculate $\lP(\ux_{n+1}=x_j\giv\uss )$ and
$\uP(\ux_{n+1}=x_j\giv\uss )$ for each $x_j\in\uc$, using a set of
priors $\mzero$ such that $\lP(\ux_{n+1}=x_j)=0$ and
$\uP(\ux_{n+1}=x_j)=1$ for each $x_j\in\uc$.

A possible near-ignorance set of priors for this problem is the set
$\mzero$ used in the IDM. We have seen, in
Section~\ref{sec:nearignorance}, that this particular near-ignorance
set of priors is such that $\lP(\ux_{n+1}=x_j)=0$ and
$\uP(\ux_{n+1}=x_j)=1$ for each $x_j\in\uc$. For this particular
choice, the following
theorem\footnote{Theorem~\ref{thm:multivacuous} is a slightly
extended version of Theorem 1 in Piatti et al. \cite{Piatti2005}.}
states necessary and sufficient conditions for learning.

\begin{thm}\label{thm:multivacuous}
Let $\Lambda (\us_i)$ be the emission matrix of $\us_i$ for
$i=1,\ldots ,n$. Let $\mzero$ be the near-ignorance set of priors
used in the IDM. Given an arbitrary observed dataset $\uss$, we
obtain a posteriori the following inferences.
\begin{itemize}
\item[1.] If all the elements of matrices $\Lambda(\us_i)$ are nonzero,
then, $\overline{P}(\ux_{n+1}=x_j\giv\uss)=1$,
$\underline{P}(\ux_{n+1}=x_j\giv\uss)=0$, for every $x_j\in\uc$.
\item[2.] $\overline{P}(\ux_{n+1}=x_j\giv\uss)<1$ for some $x_j\in\uc$, iff we observed
at least one manifest variable $\us_i=s_h$ such that
$\lambda_{hj}(\us_i)=0$.
\item[3.] $\underline{P}(\ux_{n+1}=x_j\giv\uss)>0$ for some $x_j\in\uc$, iff we observed
at least one manifest variable $\us_i=s_h$ such that
$\lambda_{hj}(\us_i)\neq 0$ and $\lambda_{hr}(\us_i)=0$ for each
$r\neq j$ in $\{1,\ldots, k\}$.
\end{itemize}
\end{thm}
In other words, to avoid vacuous posterior predictive probabilities
for the next outcome, we need at least a partial perfection of the
observational process. Some simple criteria to recognize settings
producing vacuous inferences are the following.

\begin{cor}\label{cor:multivacuous}
Under the assumptions of Theorem~\ref{thm:multivacuous}, the
following criteria hold:
\begin{enumerate}
\item[1.] If the $j$-th columns of matrices $\Lambda(\us_i)$ have all
nonzero elements, then, for each $\uss$,
$\overline{P}(\ux_{n+1}=x_j\giv\uss)=1$.
\item[2.] If the $j$-th rows of matrices $\Lambda(\us_i)$ have more than
one nonzero element, then, for each $\uss$,
$\underline{P}(\ux_{n+1}=x_j\giv\uss)=0$.
\end{enumerate}
\end{cor}

\begin{exmp}\label{exmp:diagnostictest2}\rm
Consider again the medical test of
Example~\ref{exmp:diagnostictest}. The manifest variable $\us_i$
(the result of the medical test applied to the $i$-th individual) is
a binary variable with outcomes \emph{positive} ($\postest$) or
\emph{negative} ($\negtest$). The underlying latent variable $\ux_i$
(the health status of the $i$-th individual) is also a binary
variable, with outcomes \emph{ill} ($\ill$) or \emph{healthy}
($\healthy$). The emission matrix in this case is the same for each
$i\in\enne$ and is the $2\times 2$ matrix,
\beqn
  \Lambda=\left(\begin{array}{cc} 1-\varepsilon_2 & \varepsilon_1\\
\varepsilon_2 & 1-\varepsilon_1\end{array}\right).
\eeqn
All the elements of $\Lambda$ are different from zero. Therefore,
using as set of priors the near-ignorance set of priors $\mzero$ of
the IDM, according to Theorem~\ref{thm:multivacuous}, we are unable
to move away from the initial state of ignorance. This result
confirms, in the case of the near-ignorance set of priors of the
IDM, the general result of Example~\ref{exmp:diagnostictest}.

It is interesting to remark that it is impossible to learn for
arbitrarily small values of $\varepsilon_1$ and $\varepsilon_2$,
provided that they are positive. It follows that there are
situations where the observational process cannot be neglected, even
when we deem it to be imperfect with tiny probability. This point is
particulary interesting when compared to what would be obtained
using a model with a single non-informative prior. In this case, the
difference between a model with perfect observations and a model
that takes into account the probability or error would be very small
and therefore the former model would be used instead of the latter.
Our results show that this procedure, that is almost an automatism
when using models with a single prior, may not be justified in
models with sets of priors. The point here seems to be that the
amount of imperfection of the observational process should not be
evaluated in absolute terms; it should rather be evaluated in
comparison with the weakness of the prior beliefs.
\eoe
\end{exmp}

The previous example has been concerned with the case in which the
IDM is applied to a latent categorical variable. Now we focus on the
original setup for which the IDM was conceived, where there are no
latent variables. In this case, it is well known that the IDM leads
to non-vacuous posterior predictive probabilities for the next
outcome. In the next example, we show how such a setup makes the IDM
avoid the theoretical limitations stated in
Section~\ref{sec:conditions,subsec:generalcase}.

\begin{exmp}\label{exmp:idm}\rm
In the IDM, we assume that the IID categorical variables
$(\ux_i)_{i\in\enne}$ are observable. In other words, we have
$\us_i=\ux_i$ for each $i\geq 1$ and therefore the IDM is not a
latent variable model. The IDM is equivalent to a model with
categorical manifest variables and emission matrices equal to the
identity matrix $I$. Therefore, according to the second and third
statements of Theorem~\ref{thm:multivacuous}, if $\uxs$ contains
only observations of the type $x_j$, then
\beqn
  \lP(\ux_{n+1}=x_j\giv\uxs)>0\;, \uP(\ux_{n+1}=x_j\giv\uxs)=1,
\eeqn
\beqn
  \lP(\ux_{n+1}=x_h\giv\uxs)=0\;, \uP(\ux_{n+1}=x_h\giv\uxs)<1,
\eeqn
for each $h\neq j$. Otherwise, for all the other possible observed
dataset $\uxs$,
\beqn
  \lP(\ux_{n+1}=x_j\giv\uxs)>0\;, \uP(\ux_{n+1}=x_j\giv\uxs)<1\;,
\eeqn
for each $j\in\{1,\ldots,k\}$. It follows that, in general, the IDM
produces, for each observed dataset $\uxs$, non-vacuous posterior
predictive probabilities for the next outcome.

The IDM avoids the theoretical limitations highlighted in
Section~\ref{sec:conditions,subsec:generalcase} thanks to its
particular likelihood function. Having observed
$\manifest=\realized=\uxs$, we have
\beqn
  \P(\manifest=\uxs\giv\boldsymbol{\vartheta})=\P(\realized=\uxs\giv\boldsymbol{\vartheta})=\prod_{i=1}^k
  \vartheta_i^{n_i},
\eeqn
where $n_i$ denotes the number of times that
$x_i\in\uc$ has been observed in $\uxs$. We have
$\P(\realized=\uxs\giv\boldsymbol{\vartheta})=0$ for all $\boldsymbol{\vartheta}$ such that
$\vartheta_j=0$ for at least one $j$ such that $n_j>0$ and
$\P(\realized=\uxs\giv\boldsymbol{\vartheta})>0$ for all the other
$\boldsymbol{\vartheta}\in\Theta$, in particular for all $\boldsymbol{\vartheta}$ in the interior of
$\Theta$.

Consider, to make things simpler, that in $\uxs$ at least two
different outcomes have been observed. The posterior predictive
probabilities for the next outcome are obtained calculating the
lower and upper expectations of the function
$f(\boldsymbol{\vartheta})=\vartheta_j$ for all $j\in\{1,\ldots
,k\}.$ This function reaches its minimum ($f_{\min}=0$) if
$\vartheta_j=0$ and its maximum ($f_{\min}=1$) if $\vartheta_j=1$.
Therefore, the points where the function
$f(\boldsymbol{\vartheta})=\vartheta_j$ reaches its minimum, resp.
its maximum, are on the boundary of $\ste$ and it is easy to show
that the likelihood function equals zero at least in one of these
points. It follows that the positivity assumptions of Theorem
~\ref{thm:fondamentale} are not met. \eoe
\end{exmp}

Example~\ref{exmp:idm} shows that we are able to learn, using a
near-ignorance set of priors, only if the likelihood function
$\P(\uss\giv\boldsymbol{\vartheta})$ is equal to zero in some
critical points. The likelihood function of the IDM is very
peculiar, being in general equal to zero on some parts of the
boundary of $\Theta$, and allows therefore to use a near-ignorance
set of priors $\mzero$ that models in a satisfactory way a condition
of prior (near-) ignorance.\footnote{See Walley \cite{Walley1996}
and Bernard \cite{Bernard2005} for an in-depth discussion on the
properties of the IDM.}

Yet, since the variables $(\ux_i)_{i\in\enne}$ are assumed to be
observable, the successful application of a near-ignorance set of
priors in the IDM is not helpful in addressing the doubts raised by
our theoretical results about the applicability of near-ignorance
set of priors in situations, where the variables
$(\ux_i)_{i\in\enne}$ are latent, as shown in
Example~\ref{exmp:diagnostictest2}.

\section{On modeling observable quantities}\label{sec:observables}

In this section, we discuss three alternative approaches that, at a
first sight, might seem  promising to overcome the problem of
learning under prior near-ignorance. For the sake of simplicity, we
consider the particular problem of calculating predictive
probabilities for the next outcome and a very simple setting based
on the IDM. The alternative approaches are based on trying to
predict the manifest variable rather than the latent one, thus
changing perspective with respect to the previous sections. This
change of perspective is useful to consider also because on some
occasions, e.g., when the imperfection of the observational process
is considered to be low, one may deem sufficient to focus on
predicting the manifest variable. We show, however, that the
proposed approaches eventually do not solve the mentioned learning
question, which remains therefore an open problem.

Let us introduce in detail the simple setting we are going to use.
Consider a sequence of independent and identically distributed
categorical binary latent variables $(\ux_i)_{i\in\enne}$ with
unknown chances
$\boldsymbol{\theta}=(\theta_1,\theta_2)=(\theta_1,1-\theta_1)$, and
a sequence of IID binary manifest variables $(\us_i)_{i\in\enne}$
with the same possible outcomes. Since the manifest variables are
also IID, then they can be regarded as the product of an overall
multinomial data-generating process (that includes the generation of
the latent variables as well as the observational process) with
unknown chances $\boldsymbol{\xi}:=(\xi_1,\xi_2)=(\xi_1,1-\xi_1)$.
Suppose that the emission matrix $\Lambda$ is known, constant for
each $i$ and strictly diagonally dominant, i.e.,
\beqn
  \Lambda=\left(\begin{array}{cc}1-\varepsilon_2 & \varepsilon_1\\
  \varepsilon_2 & 1-\varepsilon_1\\\end{array}\right),
\eeqn
with $\varepsilon_1,\varepsilon_2\neq 0$, $\varepsilon_1<0.5$ and
$\varepsilon_2<0.5$. This simple matrix models the case in which,
for each $i$, we are observing the outcomes of the random variable
$\ux_i$ but there is a positive probability of confounding the
actual outcome of $\ux_i$ with the other one. The random variable
$\us_i$ represents our observation, while $\ux_i$ represents the
true value. A typical example for this kind of situation is the
medical example discussed in Examples \ref{exmp:diagnostictest} and
\ref{exmp:diagnostictest2}. Suppose that we have observed
$\manifest=\uss$ and our aim is to calculate
$\lP(\ux_{n+1}=x_1\giv\uss)$ and $\uP(\ux_{n+1}=x_1\giv\uss)$.

In the previous sections we have dealt with this problem by modeling
our ignorance about the chances of $\ux_{n+1}$ with a near-ignorance
set of priors and then calculating $\lP(\ux_{n+1}=x_1\giv\uss)$ and
$\uP(\ux_{n+1}=x_1\giv\uss)$. But we already know from
Example~\ref{exmp:diagnostictest} that in this case we obtain
vacuous predictive probabilities, i.e.,
\beqn
  \lP(\ux_{n+1}=x_1\giv\uss)=0,\qquad
  \uP(\ux_{n+1}=x_1\giv\uss)=1.
\eeqn
Because this approach does not produce any useful result, one could
be tempted to modify it in order to obtain non-vacuous predictive
probabilities for the next outcome. We have identified three
possible alternative approaches that we discuss below. The basic
structure of the three approaches is identical and is based on the
idea of focusing on the manifest variables, that are observable,
instead of the latent variables. The proposed structure is the
following:
\begin{itemize}
\item specify a near-ignorance set of priors for the chances $\boldsymbol{\xi}$ of $\us_{n+1}$;
\item construct predictive probabilities for the manifest variables, i.e.,
\beqn
  \lP(\us_{n+1}=x_1\giv\uss),\qquad \uP(\us_{n+1}=x_1\giv\uss);
\eeqn
\item use the predictive probabilities calculated in the previous
point to say something about the predictive probabilities
\beqn
  \lP(\ux_{n+1}=x_1\giv\uss),\qquad \uP(\ux_{n+1}=x_1\giv\uss).
\eeqn
\end{itemize}
The three approaches differ in the specification of the
near-ignorance set of priors for $\boldsymbol{\xi}$ and on the way
$\lP(\us_{n+1}=x_1\giv\uss)$ and $\uP(\us_{n+1}=x_1\giv\uss)$ are
used to reconstruct $\lP(\ux_{n+1}=x_1\giv\uss)$ and
$\uP(\ux_{n+1}=x_1\giv\uss)$.

The first approach consists in specifying a near-ignorance set of
priors for the chances $\boldsymbol{\xi}$ taking into consideration
the fact that these chances are related to the chances
$\boldsymbol{\theta}$ through the equation
\beqn
  \xi_1=(1-\varepsilon_2)\cdot \theta_1+\varepsilon_1\cdot(1-\theta_1),
\eeqn
and therefore we have $\xi_1\in [\varepsilon_1,1-\varepsilon_2]$. A
possible way to specify correctly a near-ignorance set of priors in
this case is to consider the near-ignorance set of priors $\mzero$
of the IDM on $\boldsymbol{\theta}$, consisting of standard
$beta(s,t)$ distributions, and to substitute
\beqn
  \theta_1=\frac{\xi_1-\varepsilon_1}{1-(\varepsilon_1+\varepsilon_2)},\qquad
  d\theta_1=\frac{d\xi_1}{1-(\varepsilon_1+\varepsilon_2)},
\eeqn
into all the prior distributions in $\mzero$. We obtain thus a
near-ignorance set of priors for $\boldsymbol{\xi}$ consisting of
beta distributions scaled on the set
$[\varepsilon_1,1-\varepsilon_2]$, i.e.,
\beqn
  \xi_1\sim\frac{C}{1-(\varepsilon_1+\varepsilon_2)}
  \left(\frac{\xi_1-\varepsilon_1}{1-(\varepsilon_1+\varepsilon_2)}\right)^{st_1-1}
  \left(\frac{(1-\xi_1)-\varepsilon_2}{1-(\varepsilon_1+\varepsilon_2)}\right)^{st_2-1},
\eeqn
where $C:=\frac{\Gamma (s)}{\Gamma (st_1)\Gamma(st_2)}$. But,
scaling the distributions, we incur the same problem we have
incurred with the IDM for the latent variable. Suppose that we have
observed a dataset $\uss$ containing $n_1$ times the outcome $x_1$
and $n-n_1$ times the outcome $x_2$. The likelihood function in this
case is given by $L(\xi_1,\xi_2)=\xi_1^{n_1}\cdot
(1-\xi_1)^{(n-n_1)}$. Because $\xi_1\in
[\varepsilon_1,1-\varepsilon_2]$ the likelihood functions is always
positive and therefore the extreme distributions that are present in
the near-ignorance set of priors for $\boldsymbol{\xi}$ produce
vacuous expectations for $\xi_1$, i.e.,
$\overline{E}(\xi_1\giv\uss)=1-\varepsilon_2$ and
$\underline{E}(\xi_1\giv\uss)=\varepsilon_1$. It follows that this
approach does not solve our theoretical problem. Moreover, it
follows that the inability to learn is present under near-ignorance
even when we focus on predicting the manifest variable!

The second, more naive, approach consists in using the
near-ignorance set of priors $\mzero$ used in the standard IDM to
model ignorance about $\boldsymbol{\xi}$. In this way we are
assuming (wrongly) that $\xi_1\in [0,1]$, ignoring thus the fact
that $\xi_1\in [\varepsilon_1,1-\varepsilon_2]$ and therefore
implicitly ignoring the emission matrix $\Lambda$. Applying the
standard IDM on $\boldsymbol{\xi}$ we are able to produce
non-vacuous probabilities $\lP(\us_{n+1}=x_1\giv\uss)$ and
$\uP(\us_{n+1}=x_1\giv\uss)$. Now, because $\Lambda$ is known,
knowing the value of $P(\us_{n+1}=x_1\giv\uss)$ it is possible to
reconstruct $P(\ux_{n+1}=x_1\giv\uss)$. But this approach, that on
one hand ignores $\Lambda$ and on the other hand takes it into
consideration, is clearly wrong. For example, it can be easily shown
that it can produce probabilities outside $[0,1]$.

Finally, a third possible approach could be to neglect the existence
of the latent level and consider $\us_{n+1}$ to be the variable of
interest. Applying the standard IDM on the manifest variables we are
clearly able to produce non vacuous probabilities
$\lP(\us_{n+1}=x_1\giv\uss)$ and $\uP(\us_{n+1}=x_1\giv\uss)$ that
are then simply used instead of the probabilities
$\lP(\ux_{n+1}=x_1\giv\uss)$ and $\uP(\ux_{n+1}=x_1\giv\uss)$ in the
problem of interest. This approach is the one typically followed by
those who apply the IDM in practical problems.\footnote{See Bernard
\cite{Bernard2005} for a list of applications of the IDM.} This
approach requires the user to assume perfect observability; an
assumption that appears to be incorrect in most (if not all) real
statistical problems.  And yet this procedure, despite being wrong
or hardly justifiable from a theoretical point of view, has produced
in several applications of the IDM useful results, at least from an
empirical point of view. This paradox between our theoretical
results and the current practice is an open problem that deserves to
be investigated in further research.

\section{Conclusions}\label{sec:conclusions}

In this paper we have proved a sufficient condition that prevents
learning about a latent categorical variable to take place under
prior near-ignorance regarding the data-generating process.

The condition holds as soon as the likelihood is strictly positive
(and continuous), and so is satisfied frequently, even in the more
common and simple settings. Taking into account that the considered
framework is very general and pervasive of statistical practice, we
regard this result as a form of strong evidence against the
possibility to use prior near-ignorance in real statistical
problems. Given also that prior near-ignorance is arguably a
privileged way to model a state of ignorance, our results appear to
substantially reduce the hope to be able to adopt a form of prior
ignorance to do objective-minded statistical inference.

With respect to future research, two possible research directions
seem to be particularly important to investigate.

As reported by Bernard \cite{Bernard2005}, near-ignorance sets of
priors, in the specific form of the IDM, have been successfully used
in a number of applications. On the other hand, the theoretical
results presented in this paper point to the impossibility of
learning in real statistical problems when starting from a state of
near-ignorance. This paradox between empirical and theoretical
results should be investigated in order to better understand the
practical relevance of the theoretical analysis presented here, and
more generally to explain the mechanism behind such an apparent
contradiction.

The proofs contained in this paper suggest that the impossibility of
learning under prior near-ignorance with latent variables is mainly
due to the presence, in the set of priors, of extreme distributions
arbitrarily close to the deterministic ones. Some preliminary
experimental analyses have shown that learning is possible as soon
as one restricts the set of priors so as to rule out the extreme
distributions. This can be realized by defining a notion of distance
between priors and then by allowing a distribution to enter the
prior set of probability distributions only if it is at least a
certain positive distance away from the deterministic priors. The
minimal distance can be chosen arbitrarily small (while remaining
positive), and this allows one to model a state of very weak
beliefs, close to near-ignorance. Such a weak state of beliefs could
keep some of the advantages of near-ignorance (although it would
clearly not be a model of ignorance) while permitting learning to
take place. The main problem of this approach is the justification,
i.e., the interpretation of the (arbitrary) restriction of the
near-ignorance set of priors. A way to address this issue might be
to identify a set of desirable principles, possibly similar to the
symmetry and embedding principles, leading in a natural way to a
suitably large set of priors describing a state close to
near-ignorance.
\section*{Acknowledgements}
This work was partially supported by Swiss NSF grants
200021-113820/1, 200020-116674/1, 100012-105745/1, and by the Hasler
foundation grant 2233.

\begin{appendix}

\section{Technical preliminaries}\label{techresults}

In this appendix we prove some technical results that are used to
prove the theorems in the paper. First of all, we introduce some
notation used in this appendix. Consider a sequence of probability
densities $\seqpn$ and a function $f$ defined on a set $\ste$. Then
we use the notation
\beqn
  \E_n(f):=\int_{\ste} f(\boldsymbol{\vartheta}) p_n(\boldsymbol{\vartheta}) d\boldsymbol{\vartheta},\quad\P_n(\widetilde{\ste})
  :=\int_{\widetilde{\ste}} p_n(\boldsymbol{\vartheta}) d\boldsymbol{\vartheta},\quad
  \widetilde{\ste}\subseteq\ste,
\eeqn
and with $\rightarrow$ we denote $\lim_{n\rightarrow\infty}$.

\begin{thm}\label{thm:marcus1}
Let $\ste\subset\erre^k$ be the closed $k$-dimensional simplex and
let $\seqpn$ be a sequence of probability densities defined on
$\ste$ w.r.t. the Lebesgue measure. Let $f\geq 0$ be a bounded
continuous function on $\ste$ and let $f_{\max}:=\sup_{\ste}(f)$ and
$f_{\min}:=\inf_{\ste}(f)$. For this function define the measurable
sets
\beq
  \ste_{\delta}=\{\boldsymbol{\vartheta}\in\ste\giv f(\boldsymbol{\vartheta})\geq
  f_{\max}-\delta\},\label{deftetadelta}
\eeq
  \beq\widetilde{\ste}_{\delta}=\{\boldsymbol{\vartheta}\in\ste\giv f(\boldsymbol{\vartheta})\leq
  f_{\min}+\delta\}.\label{deftetadeltainf}
\eeq
\begin{enumerate}
\item Assume that $\seqpn$ concentrates on a maximum of $f$ for
$n\rightarrow\infty$, in the sense that
\beq
  \E_n(f)\rightarrow f_{\max},\label{convfmax}
\eeq
then, for all $\delta >0$, it holds
\beqn
  \P_n(\ste_{\delta})\rightarrow 1.
\eeqn
\item Assume that $\seqpn$ concentrates on a minimum of $f$ for
$n\rightarrow\infty$, in the sense that
\beq
  \E_n(f)\rightarrow f_{\min},\label{convfmin}
\eeq
then, for all $\delta >0$, it holds
\beqn
  \P_n(\widetilde{\ste}_{\delta})\rightarrow 1.
\eeqn
\end{enumerate}
\end{thm}

\noindent{\bf Proof.} We begin by proving the first statement. Let
$\delta>0$ be arbitrary and
$\bar\ste_{\delta}:=\ste\setminus\ste_{\delta}$. From
(\ref{deftetadelta}) we know that on $\ste_{\delta}$ it holds
$f(\boldsymbol{\vartheta})\geq f_{\max}-\delta$,  and therefore on
$\bar\ste_{\delta}$ we have $f(\boldsymbol{\vartheta})\leq
f_{\max}-\delta$, and thus
\beq
  \frac{f_{\max}-f(\boldsymbol{\vartheta})}{\delta}\geq 1. \label{fmaxmenofmagg1}
\eeq
It follows that
\bqan
  1-\P_n(\ste_{\delta}) & = &
  \P_n(\bar\ste_{\delta})=\int_{\bar\ste_{\delta}} p_n(\boldsymbol{\vartheta})
  d\boldsymbol{\vartheta}\stackrel{(\ref{fmaxmenofmagg1})}{\leq}
  \int_{\bar\ste_{\delta}}
  \frac{f_{\max}-f(\boldsymbol{\vartheta})}{\delta} p_n(\boldsymbol{\vartheta}) d\boldsymbol{\vartheta}\\
  & \leq & \int_{\ste} \frac{f_{\max}-f(\boldsymbol{\vartheta})}{\delta} p_n(\boldsymbol{\vartheta})
  d\boldsymbol{\vartheta}=\frac{1}{\delta}(f_{\max}-\E_n(f))\stackrel{(\ref{convfmax})}{\longrightarrow}
  0,
\eqan
and therefore $\P_n(\ste_{\delta})\rightarrow 1$ and thus the first
statement is proved. To prove the second statement, let $\delta>0$
be arbitrary and
$\widehat{\ste}_{\delta}:=\ste\setminus\widetilde{\ste}_{\delta}$.
From (\ref{deftetadeltainf}) we know that on
$\widetilde{\ste}_{\delta}$ it holds $f(\boldsymbol{\vartheta})\leq
f_{\min}+\delta$, and therefore on $\widehat{\ste}_{\delta}$ we have
$f(\boldsymbol{\vartheta})\geq f_{\min}+\delta$, and thus
\beq
  \frac{f(\boldsymbol{\vartheta})-f_{\min}}{\delta}\geq 1. \label{fmenofminmagg1}
\eeq
It follows that
\bqan
  1-\P_n(\widetilde{\ste}_{\delta}) & = &
  \P_n(\widehat{\ste}_{\delta})=\int_{\widehat{\ste}_{\delta}}
  p_n(\boldsymbol{\vartheta}) d\boldsymbol{\vartheta}\stackrel{(\ref{fmenofminmagg1})}{\leq}
  \int_{\widehat{\ste}_{\delta}}
  \frac{f(\boldsymbol{\vartheta})-f_{\min}}{\delta} p_n(\boldsymbol{\vartheta}) d\boldsymbol{\vartheta}\\
  & \leq & \int_{\ste} \frac{f(\boldsymbol{\vartheta})-f_{\min}}{\delta} p_n(\boldsymbol{\vartheta})
  d\boldsymbol{\vartheta}=\frac{1}{\delta}(\E_n(f)-f_{\min})\stackrel{(\ref{convfmax})}{\longrightarrow}
  0,
\eqan
and therefore $\P_n(\widetilde{\ste}_{\delta})\rightarrow 1$. \qed

\begin{thm}\label{thm:marcus2}
Let $L(\boldsymbol{\vartheta})\geq 0$ be a bounded measurable function and suppose
that the assumptions of Theorem~\ref{thm:marcus1} hold. Then the
following two statements hold.
\begin{enumerate}
\item If the function $L(\boldsymbol{\vartheta})$ is such that
\beq
  c:=\lim_{\delta\rightarrow 0} \inf_{\boldsymbol{\vartheta}\in\ste_{\delta}} L(\boldsymbol{\vartheta})>0, \label{liminfpos}
\eeq
and $\seqpn$ concentrates on a maximum of $f$ for
$n\rightarrow\infty$, then
\beq
  \frac{\E_n(Lf)}{\E_n(L)}=\frac{\int_{\ste} f(\boldsymbol{\vartheta}) L(\boldsymbol{\vartheta})
  p_n(\boldsymbol{\vartheta}) d\boldsymbol{\vartheta}}{\int_{\ste} L(\boldsymbol{\vartheta}) p_n(\boldsymbol{\vartheta})
  d\boldsymbol{\vartheta}}\rightarrow f_{\max}. \label{vacuousthm2}
\eeq
\item If the function $L(\boldsymbol{\vartheta})$ is such that
\beq
  c:=\lim_{\delta\rightarrow 0} \inf_{\boldsymbol{\vartheta}\in\widetilde{\ste}_{\delta}}
  L(\boldsymbol{\vartheta})>0, \label{liminfposinf}
\eeq
and $\seqpn$ concentrates on a minimum of $f$ for
$n\rightarrow\infty$, then
\beq
  \frac{\E_n(Lf)}{\E_n(L)}\longrightarrow f_{\min}.
  \label{vacuousthm2inf}
\eeq
\end{enumerate}
\end{thm}

\begin{rem}\label{rem:assumptionmarcus}
If $L$ is strictly positive in each point in $\ste$ where the
function $f$ reaches its maximum, resp. minimum, and is continuous
in an arbitrary small neighborhood of those points, then
(\ref{liminfpos}), resp. (\ref{liminfposinf}), are satisfied.
\end{rem}

\noindent{\bf Proof.} We begin by proving the first statement of the
theorem. Fix $\varepsilon$ and $\delta$ arbitrarily small, but
$\delta$ small enough such that
$\inf_{\boldsymbol{\vartheta}\in\ste_{\delta}}
L(\boldsymbol{\vartheta})\geq \frac{c}{2}$. denote by $L_{\max}$ the
supremum of the function $L(\boldsymbol{\vartheta})$ in $\ste$. From
Theorem~\ref{thm:marcus1}, we know that $\P_n(\ste_{\delta})\geq
1-\varepsilon$, for $n$ sufficiently large. This implies, for $n$
sufficiently large,
\beq
  \label{thmvacue3} \E_n(L)=\int_{\ste} L(\boldsymbol{\vartheta}) p_n(\boldsymbol{\vartheta}) d\boldsymbol{\vartheta}
  \geq \int_{\ste_{\delta}} L(\boldsymbol{\vartheta}) p_n(\boldsymbol{\vartheta}) d\boldsymbol{\vartheta}\geq
  \frac{c}{2} (1-\varepsilon),
\eeq
\beq
  \label{thmvacue1} \E_n(Lf)\leq\E_n(Lf_{\max}) = f_{\max} \E_n(L),
\eeq
\bqa
  \E_n(L) & = &\int_{\bar\ste_{\delta}} L(\boldsymbol{\vartheta}) p_n(\boldsymbol{\vartheta}) d\boldsymbol{\vartheta}
  +
  \int_{\ste_{\delta}} L(\boldsymbol{\vartheta}) p_n(\boldsymbol{\vartheta}) d\boldsymbol{\vartheta}\nonumber\\
  & \leq & L_{\max} \int_{\bar\ste_{\delta}} p_n(\boldsymbol{\vartheta})
  d\boldsymbol{\vartheta} + \int_{\ste_{\delta}}
  \underbrace{\frac{f(\boldsymbol{\vartheta})}{f_{\max}-\delta}}_{\geq
  1\,\mathrm{\small{on}}\,\ste_{\delta}} L(\boldsymbol{\vartheta}) p_n(\boldsymbol{\vartheta}) d\boldsymbol{\vartheta}
  \nonumber\\& \leq & L_{\max}\cdot
  \varepsilon+\frac{1}{f_{\max}-\delta}
  \E_n(Lf).\label{thmvacue2}\\\nonumber
\eqa
Combining (\ref{thmvacue3}), (\ref{thmvacue1}) and
(\ref{thmvacue2}), we have
\beq
  f_{\max} \geq \frac{\E_n(Lf)}{\E_n(L)}\geq
  (f_{\max}-\delta)\frac{\E_n(L)-L_{\max}\cdot\varepsilon}{\E_n(L)}\geq
  (f_{\max}-\delta) \left(
  1-\frac{L_{\max}\cdot\varepsilon}{\frac{c}{2}(1-\varepsilon)}\right).
  \nonumber
\eeq
Since the right-hand side of the last inequality tends to $f_{\max}$
for $\delta,\varepsilon\rightarrow 0$, and both $\delta,\varepsilon$
can be chosen arbitrarily small, we have
\beqn
  \frac{\E_n(Lf)}{\E_n(L)}\rightarrow f_{\max}.
\eeqn
To prove the second statement of the theorem, fix $\varepsilon$ and
$\delta$ arbitrarily small, but $\delta$ small enough such that
$\inf_{\boldsymbol{\vartheta}\in\widetilde{\ste}_{\delta}} L(\boldsymbol{\vartheta})\geq
\frac{c}{2}$. From Theorem~\ref{thm:marcus1}, we know that
$\P_n(\widetilde{\ste}_{\delta})\geq 1-\varepsilon$, for $n$
sufficiently large and therefore $\P_n(\widehat{\ste}_{\delta})\leq
\varepsilon$. This implies, for $n$ sufficiently large,
\beq
  \label{thmvacue3inf} \E_n(L)=\int_{\ste} L(\boldsymbol{\vartheta}) p_n(\boldsymbol{\vartheta})
  d\boldsymbol{\vartheta} \geq \int_{\widetilde{\ste}_{\delta}} L(\boldsymbol{\vartheta}) p_n(\boldsymbol{\vartheta})
  d\boldsymbol{\vartheta}\geq \frac{c}{2} (1-\varepsilon),
\eeq
\beq
  \label{thmvacue1inf} \E_n(Lf)\geq\E_n(Lf_{\min}) = f_{\min}
  \E_n(L)\Rightarrow f_{\min}\leq \frac{\E_n(Lf)}{\E_n(L)}.
\eeq
Define the function
\beqn
  K(\boldsymbol{\vartheta}):=\left(1-\frac{f(\boldsymbol{\vartheta})}{f_{\min}+\delta}\right)L(\boldsymbol{\vartheta}).
\eeqn
By definition, the function $K$ is negative on
$\widehat{\ste}_{\delta}$ and is bounded. denote by $K_{\min}$ the
(negative) minimum of $K$. We have
\bqan
  \E_n(L) & = & \int_{\widehat{\ste}_{\delta}} L(\boldsymbol{\vartheta}) p_n(\boldsymbol{\vartheta})
  d\boldsymbol{\vartheta}+\int_{\widetilde{\ste}_{\delta}} L(\boldsymbol{\vartheta}) p_n(\boldsymbol{\vartheta}) d\boldsymbol{\vartheta}\\
  & \geq & \int_{\widehat{\ste}_{\delta}} L(\boldsymbol{\vartheta}) p_n(\boldsymbol{\vartheta})
  d\boldsymbol{\vartheta}+\int_{\widetilde{\ste}_{\delta}}
  \underbrace{\frac{f(\boldsymbol{\vartheta})}
  {f_{\min}+\delta}}_{\leq 1\,\textrm{on}\,\widetilde{\ste}_{\delta}}L(\boldsymbol{\vartheta}) p_n(\boldsymbol{\vartheta}) d\boldsymbol{\vartheta}\\
  & = & \int_{\widehat{\ste}_{\delta}}
  \underbrace{\left(L(\boldsymbol{\vartheta})-\frac{f(\boldsymbol{\vartheta})}
  {f_{\min}+\delta}L(\boldsymbol{\vartheta})\right)}_{=K(\boldsymbol{\vartheta})} p_n(\boldsymbol{\vartheta})
  d\boldsymbol{\vartheta}+\frac{1} {f_{\min}+\delta}\underbrace{\int_{\ste}
  f(\boldsymbol{\vartheta}) L(\boldsymbol{\vartheta}) p_n(\boldsymbol{\vartheta}) d\boldsymbol{\vartheta}}_{=\E_n(Lf)}\\
  & \geq & K_{\min}\cdot\P_n(\widehat{\ste}_{\delta})+\frac{1}
  {f_{\min}+\delta}\cdot\E_n(Lf).\\
\eqan
It follows that
\beqn
  \left(\E_n(L)-K_{\min}\cdot\P_n(\widehat{\ste}_{\delta})\right)(f_{\min}+\delta)\geq\E_n(Lf),
\eeqn
and thus, combining the last inequality with (\ref{thmvacue3inf})
and (\ref{thmvacue1inf}), we obtain
\bqan
  f_{\min}\leq\frac{\E_n(Lf)}{\E_n(L)} & \leq &
  (f_{\min}+\delta)\left(1+\frac{|K_{\min}|\cdot
  P_n(\widehat{\ste}_{\delta})}{\E_n(L)}\right)\\
  & \leq & (f_{\min}+\delta)\left(1+\frac{|K_{\min}|\cdot
  \varepsilon}{\frac{c}{2}(1-\varepsilon)}\right).\\
\eqan
Since the right-hand side of the last inequality tends to $f_{\min}$
for $\delta,\varepsilon\rightarrow 0$, and both $\delta,\varepsilon$
can be chosen arbitrarily small, we have
\beqn
  \frac{\E_n(Lf)}{\E_n(L)}\rightarrow f_{\min}.
\vspace*{-4ex}\eeqn
\qed

\section{Proofs of the main results}

\noindent{\bf Proof of Theorem~\ref{thm:fondamentale}.}
Define, $f_{\min}:=\inf_{\boldsymbol{\vartheta}\in\ste} f(\boldsymbol{\vartheta})$, $
f_{\max}:=\sup_{\boldsymbol{\vartheta}\in\ste} f(\boldsymbol{\vartheta}),$ and define the
bounded non-negative function
$\tilde{f}(\boldsymbol{\vartheta}):=f(\boldsymbol{\vartheta})-f_{\min}\geq 0$. We have,
$\tilde{f}_{\max}=f_{\max}-f_{\min}$. If $\mzero$ is such that a
priori, $\uE(f)=f_{\max}$, then we have also that
$\uE(\tilde{f})=\tilde{f}_{\max}$, because,
\beqn
  \uE(\tilde{f})=\sup_{p\in\mzero} E_p(f-f_{\min})=\sup_{p\in\mzero}
  E_p(f)-f_{\min}=\uE(f)-f_{\min}=f_{\max}-f_{\min}=\tilde{f}_{\max}.
\eeqn
Then, it is possible to define a sequence
$(p_n)_{n\in\enne}\subset\mzero$ such that
$\E_n(\tilde{f})\rightarrow \tilde{f}_{\max}$. According to
Theorem~\ref{thm:marcus2}, substituting $L(\boldsymbol{\vartheta})$ with
$\P(\uss\giv\boldsymbol{\vartheta})$ in (\ref{vacuousthm2}), we see that
$\E_n(\tilde{f}\giv\uss)\rightarrow \tilde{f}_{\max}=\uE(\tilde{f})$
and therefore $\uE(\tilde{f}\giv\uss)=\uE(\tilde{f})$, from which
follows that,
\beqn
  \uE(f\giv\uss)-f_{\min}=\uE(f)-f_{\min}=f_{\max}-f_{\min}.
\eeqn
We can conclude that, $\uE(f\giv\uss)=\uE(f)=f_{\max}.$ In the same
way, substituting $\lE$ to $\uE$, we can prove that
$\lE(f\giv\uss)=\lE(f)=f_{\min}.$
\qed

Corollary~\ref{cor:fondamentale} is a direct consequence of
Theorem~\ref{thm:fondamentale}.

\vspace{1ex}\noindent{\bf Proof of Theorem~\ref{thm:multivacuous}}
To prove Theorem~\ref{thm:multivacuous} we need the following lemma.
\begin{lem}\label{lem:lemma1} Consider a dataset $\uxs$ with frequencies $\countsx$. Then, the following equality
holds,
\beqn
  \prod_{h=1}^k \vartheta_h^{a_h^{\uxs}}\cdot
  dir_{s,\ut}(\boldsymbol{\vartheta})=\frac{\prod_{h=1}^k\cdot\prod_{j=1}^{a_h^{\uxs}}(st_h+j-1)}{\prod_{j=1}^n
  (s+j-1)}\cdot dir_{s^{\uxs},\ut^{\uxs}}(\boldsymbol{\vartheta}),
\eeqn
where $s^{\uxs}:=n+s$ and $t^{\uxs}_h:=\frac{a_h^{\uxs}+s
t_h}{n+s}$. When $a_h^{\uxs}=0$, we set
$\prod_{j=1}^{0}(st_h+j-1):=1$ by definition. \label{propo}
\end{lem}
A proof of Lemma~\ref{lem:lemma1} is in \cite{Piatti2005}. Because
$\P(\uxs\giv\boldsymbol{\vartheta})=\prod_{h=1}^k \vartheta_h^{a_h^{\uxs}}$, according
to Bayes' rule, we have
$p(\boldsymbol{\vartheta}\giv\uxs)=dir_{s^{\uxs},\ut^{\uxs}}(\boldsymbol{\vartheta})$ and
\beq
  P(\uxs)=\frac{\prod_{h=1}^k
  \prod_{l=1}^{a_h^{\uxs}}(st_h+l-1)}{\prod_{l=1}^n
  (s+l-1)}.\label{probx}
\eeq
Given a Dirichlet distribution
$dir_{s,\ut}(\boldsymbol{\vartheta})$, the expected value
$\E(\vartheta_j)$ is given by $\E(\vartheta_j)=t_j$ (see
\cite{Kotz}). It follows that
\beqn
  \E(\vartheta_j\giv\uxs)=\ut_j^{\uxs}=\frac{a_j^{\uxs}+s t_j}{n+s}.
\eeqn
We are now ready to prove Theorem~\ref{thm:multivacuous}.
\begin{itemize}
\item[1.] The first statement of Theorem~\ref{thm:multivacuous} is a consequence of Corollary~\ref{cor:fondamentale4}.
Because $\us_i$ is independent of $\boldsymbol{\vartheta}$ given $\ux_i$ for each
$i\in\enne$, we have
\beq
  \P(\uss\giv\uxs,\boldsymbol{\vartheta})=\P(\uss\giv\uxs),\label{independence}
\eeq
and therefore, using (\ref{independence}) and Bayes' rule, we obtain
the likelihood function,
\beq
  L(\boldsymbol{\vartheta})=P(\uss\giv\boldsymbol{\vartheta})=\sum_{\uxs\in\ucn} P(\uss\giv\uxs)\cdot
  P(\uxs\giv\boldsymbol{\vartheta})=\sum_{\uxs\in\ucn} P(\uss\giv\uxs)\cdot
  \prod_{h=1}^k \vartheta_h^{a_h^{\uxs}}. \label{likelihoodHIDM}
\eeq
Because all the elements of the matrices $\Lambda^{S_i}$ are
nonzero, we have $P(\uss\giv\uxs)>0$, for each $\uss$ and each
$\uxs\in\ucn$. For each $\boldsymbol{\vartheta}\in\ste$, there is at least one
$\uxs\in\ucn$ such that $\prod_{h=1}^k \vartheta_h^{a_h^{\uxs}}>0$.
It follows that,
\beqn
  L(\boldsymbol{\vartheta})=\sum_{\uxs\in\ucn}
  P(\uss\giv\uxs)\cdot \prod_{j=1}^k \vartheta_j^{a_j^{\uxs}}>0,
\eeqn
for each $\boldsymbol{\vartheta}\in\ste$ and therefore, according to
Corollary ~\ref{cor:fondamentale4} with $n'=1$, the predictive
probabilities that are vacuous a priori remain vacuous also a
posteriori.

\item[2.] We have $\P(\ux_{n+1}=x_j\giv\uss)=\E(\vartheta_j\giv\uss)$,
and therefore, according to Lemma~\ref{lem:lemma1} and Bayes' rule,
\bqa
  P(\ux_{n+1}=x_j\giv\uss) & = & \frac{\int_{\ste} \vartheta_j
  P(\uss\giv\boldsymbol{\vartheta}) p(\boldsymbol{\vartheta}) d\boldsymbol{\vartheta}}{\int_{\ste}
  P(\uss\giv\boldsymbol{\vartheta}) p(\boldsymbol{\vartheta}) d\boldsymbol{\vartheta}}=\nonumber\\
  & \stackrel{(\ref{independence})}= &
  \frac{\sum_{\uxs\in\ucn}\int_{\ste} \vartheta_j P(\uss\giv\uxs)
  P(\uxs\giv\boldsymbol{\vartheta}) p(\boldsymbol{\vartheta}) d\boldsymbol{\vartheta}}{\sum_{\uxs\in\ucn}\int_{\ste}
  P(\uss\giv\uxs) P(\uxs\giv\boldsymbol{\vartheta}) p(\boldsymbol{\vartheta}) d\boldsymbol{\vartheta}}=\nonumber\\ & =
  & \frac{\sum_{\uxs\in\ucn} P(\uss\giv\uxs) P(\uxs) \int_{\ste}
  \vartheta_j p(\boldsymbol{\vartheta}\giv\uxs) d\boldsymbol{\vartheta}}{\sum_{\uxs\in\ucn}
  P(\uss\giv\uxs)
  P(\uxs)}=\nonumber\\
  & = & \sum_{\uxs\in\ucn}\left( \frac{P(\uss\giv\uxs)
  P(\uxs)}{\sum_{\uxs\in\ucn} P(\uss\giv\uxs)
  P(\uxs)}\right)\cdot\E(\vartheta_j\giv\uxs),\nonumber\\& = &
  \sum_{\uxs\in\ucn}\left( \frac{P(\uss\giv\uxs)
  P(\uxs)}{\sum_{\uxs\in\ucn} P(\uss\giv\uxs)
  P(\uxs)}\right)\cdot\frac{a_j^{\uxs}+st_j}{n+s}.\label{postprobs}\\\nonumber
\eqa
It can be checked that the denominator of (\ref{postprobs}) is
positive and therefore conditioning on events with zero probability
is not a problem in this setting. (\ref{postprobs}) is a convex sum
of fractions and is therefore a continuous function of $t$ on
$\sett$. Denote by $\overline{\uxs}^j$ the dataset of length $n$
composed only by outcomes $x_j$, i.e., the dataset with
$a_j^{\overline{\uxs}^j}=n$ and $a_h^{\overline{\uxs}^j}=0$ for each
$h\neq j$. For all $\uxs\neq\overline{\uxs}^j$ we have
\beqn
  \frac{a_j^{\uxs}+st_j}{n+s}\leq \frac{n-1+st_j}{n+s}\leq
  \frac{n-1+s}{n+s}<1,
\eeqn
on $\overline{\sett}$ (the closure of $\sett$), only
$\overline{\uxs}^j$ has
\beqn
  \sup_{t\in\sett}
  \frac{a_j^{\overline{\uxs}^j}+st_j}{n+s}=\sup_{t\in\sett}
  \frac{n+st_j}{n+s}=1.
\eeqn
A convex sum of fractions smaller than or equal to one is equal to
one, only if the weights associated to fractions smaller than one
are all equal to zero and there are some positive weights associated
to fractions equal to one. If $P(\uss\giv\overline{\uxs}^j)=0$, then
(\ref{postprobs}) is a convex combination of fractions strictly
smaller than 1 on $\overline{\sett}$ and therefore
$\overline{P}(\ux_{n+1}=x_j\giv\uss)<1$. If
$P(\uss\giv\overline{\uxs}^j)\neq 0$, then letting $t_j\rightarrow
1$, and consequently $t_h\rightarrow 0$ for all $h\neq j$, according
to (\ref{probx}), we have $P(\overline{\uxs}^j)\rightarrow 1$ and
$P(\uxs)\rightarrow 0$ for all $\uxs\neq \overline{\uxs}^j$, and
thus, using (\ref{postprobs}),
\beqn
  1\geq\overline{P}(\ux_{n+1}=x_j\giv\uss)\geq\lim_{t_j\rightarrow 1}
  P(\ux_{n+1}=x_j\giv\uss)=\frac{P(\uss\giv\overline{\uxs}^j)
  P(\overline{\uxs}^j)\frac{n+s}{n+s}}{P(\uss\giv\overline{\uxs}^j)P(\overline{\uxs}^j)}=1.
\eeqn
If we have observed a manifest variable $\us_i=s_h$ with
$\lambda_{hj}^{\us_t}=0$, it means that the observation excludes the
possibility that the underlying value of $\ux_i$ is $x_j$, therefore
$P(\uss\giv \overline{\uxs}^j)=0$ and thus
\beqn
  \overline{P}(\ux_{n+1}=x_j\giv\uss)<1.
\eeqn
On the other hand, if $\overline{P}(\ux_{n+1}=x_j\giv\uss)<1$, it
must hold that $P(\uss\giv \overline{\uxs}^j)=0$, i.e., that we have
observed a realization of a manifest that is incompatible with the
underlying (latent) outcome $x_j$. But a realization of a manifest
that is incompatible with the underlying (latent) outcome only if
the observed manifest variable was $\us_i=s_h$ with
$\lambda_{hj}^{\us_i}=0$.

\item[3.] Having observed a manifest variable $\us_i=s_h$,
such that $\lambda_{hj}^{\us_i}\neq 0$ and $\lambda_{hr}^{\us_i}=
0$ for each $r\neq j$ in $\{1,\ldots ,k\}$, we are sure that the
underlying value of $\ux_i$ is $x_j$. Therefore, $P(\uss\giv\uxs)=0$
for all $\uxs$ with $a_j^{\uxs}=0$. It follows from
(\ref{postprobs}) that
\beqn
  P(\ux_{n+1}=x_j\giv\uss)=\frac{\sum_{\uxs\in\ucn,\: a_j^{\uxs}>0}
  P(\uss\giv\uxs)
  P(\uxs)\cdot\frac{a_j^{\uxs}+st_j}{n+s}}{\sum_{\uxs\in\ucn,\:
  a_j^{\uxs}>0} P(\uss\giv\uxs) P(\uxs)},
\eeqn
which is a convex
combination of terms
\beqn
  \frac{a_j^{\uxs}+st_j}{n+s}\geq \frac{a_j^{\uxs}}{n+s}\geq \frac{1}{n+s},
\eeqn
and is therefore
greater than zero for each $t\in\overline{\sett}$. It follows that
\beqn
  \underline{P}(\ux_{n+1}=x_j\giv\uss)\geq\frac{1}{n+s}>0.
\eeqn
On the other hand, if we do not observe a manifest variable as described above,
it exists surely at least one $\uxs$ with $a_j^{\uxs}=0$ and
$P(\uss\giv\uxs)>0$. In this case, using (\ref{postprobs}) and
letting $t_j\rightarrow 0$, we have, because of (\ref{probx}), that
$P(\uxs)\rightarrow 0$ for all $\uxs$ with $a_j^{\uxs}>0$. It
follows that
\beqn
  \lim_{t_j\rightarrow 0} P(\ux=x_j\giv\uss)=\lim_{t_j\rightarrow 0} \frac{\sum_{\uxs\in\ucn,\: a_j^{\uxs}=0}
  P(\uss\giv\uxs)
  P(\uxs)\cdot\frac{a_j^{\uxs}+st_j}{n+s}}{\sum_{\uxs\in\ucn,\:
  a_j^{\uxs}=0} P(\uss\giv\uxs) P(\uxs)}.
\eeqn
Assume for simplicity
that, for all $h\neq j$, $t_h\not\rightarrow 0$, then $P(\uxs)>0$
for all $\uxs$ with $a_j^{\uxs}=0$ and $P(\uxs)\not\rightarrow 0$.
Because, with $a_j^{\uxs}=0$, we have
\beqn
  \lim_{t_j\rightarrow 0}\frac{a_j^{\uxs}+st_j}{n+s}=\lim_{t_j\rightarrow 0}\frac{0+st_i}{n+s}=0,
\eeqn
we
obtain directly,
\beqn
  0\leq\underline{P}
  (\ux_{n+1}=x_j\giv\uss)=\inf_{t\in\sett}
  P(\ux_{n+1}=x_j\giv\uss)\leq\lim_{t_j\rightarrow 0}
  P(\ux_{n+1}=x_j\giv\uss)=0.
\vspace*{-4ex}\eeqn
\end{itemize}
\qed

Corollary~\ref{cor:multivacuous} is a direct consequence of
Theorem~\ref{thm:multivacuous}.
\end{appendix}

\begin{small}

\end{small}

\end{document}